\documentclass{article}

\PassOptionsToPackage{round}{natbib}
\usepackage[preprint]{neurips_2023}

\usepackage{natbib}
\usepackage[utf8]{inputenc} % allow utf-8 input
\usepackage[T1]{fontenc}    % use 8-bit T1 fonts
\usepackage{url}            % simple URL typesetting
\usepackage{booktabs}       % professional-quality tables
\usepackage{amsfonts}       % blackboard math symbols
\usepackage{nicefrac}       % compact symbols for 1/2, etc.
\usepackage{microtype}      % microtypography
\usepackage{xcolor}         % colors
\usepackage{booktabs} 
\usepackage{graphicx}
 \usepackage[T1]{fontenc}
\RequirePackage{fix-cm}
\let\accentvec\vec
\makeatletter
\let\@fnsymbol\@arabic
\makeatother

\usepackage[final]{pdfpages}

\usepackage{epsfig}
\usepackage{amssymb}% http://ctan.org/pkg/amssymb
\usepackage{bm}
\usepackage{pifont}% http://ctan.org/pkg/pifont
\usepackage[misc]{ifsym}
\usepackage{setspace}
\usepackage{epstopdf}
\usepackage{secdot}
\usepackage{cancel}

\let\vec\accentvec
\usepackage{amsmath}
\usepackage{mathtools}
\usepackage{dsfont}
\usepackage{siunitx}
\usepackage{latexsym}
\usepackage{mathrsfs}
\usepackage{color}
\usepackage{soul}
\usepackage{setspace}
\usepackage{parskip}

\usepackage[ruled, noend]{algorithm2e}
\usepackage[colorlinks = true, pdfstartview = FitV, linkcolor = gray, citecolor = darkgray, urlcolor = blue, bookmarks = false]{hyperref}
\newcommand\deldel[2][]{\ensuremath{\frac{\partial#1}{\partial#2}}} 
\usepackage{microtype}
\usepackage{graphicx}
\usepackage{subfigure}
\usepackage{booktabs} % for professional tables
\usepackage{amsmath, amsfonts}

\newcommand{\m}[1]{\mathrm{#1} }
\renewcommand{\cal}[1]{\mathcal{#1}}
\renewcommand{\v}[1]{\boldsymbol{#1}}
\newcommand{\bb}[1]{\mathbb{#1}}

\usepackage{xcolor, color}
\definecolor{red}{rgb}{0.81, 0.09, 0.13}

\usepackage{amsmath}
\usepackage{amssymb}
\usepackage{mathtools}
\usepackage{amsthm}

\usepackage[capitalize,noabbrev]{cleveref}
\theoremstyle{plain}
\renewcommand{\refname}{References}

\title{Federated Variational Inference Methods for Structured Latent Variable Models}

\author{%
  Conor Hassan \\
  Centre for Data Science\\
  Queensland University of Technology\\
  \texttt{conordaniel.hassan@hdr.qut.edu.au} 
  \And
  Robert Salomone \\
  Centre for Data Science \\
  Queensland University of Technology \\
  \texttt{robert.salomone@qut.edu.au}
  \AND
  Kerrie Mengersen \\
  Centre for Data Science \\
  Queensland University of Technology \\
  \texttt{k.mengersen@qut.edu.au}
}

\begin{document}

\maketitle
%\begin{bibunit}

\begin{abstract}
Federated learning methods enable model training across distributed data sources without data leaving their original locations and have gained increasing interest in various fields. However, existing approaches are limited, excluding many structured probabilistic models. We present a general and elegant solution based on structured variational inference, widely used in Bayesian machine learning, adapted for the federated setting. Additionally, we provide a communication-efficient variant analogous to the canonical FedAvg algorithm. The proposed algorithms' effectiveness is demonstrated, and their performance is compared with hierarchical Bayesian neural networks and topic models. 
\end{abstract}

\section{Introduction}\label{section:introduction}
\emph{Federated learning} (FL) is an algorithm class that allows for distributed machine learning across multiple data sources without requiring aggregation of data in a single location \citep{mcmahan2017communication, kairouz2021advances}. FL approaches involve multiple communication rounds of information between respective data sources, referred to herein as \emph{silos}, and a centralized \emph{server}. FL has significant potential for real-world applications that require secure, privacy-preserving models such as in healthcare \citep{rieke2020future, xu2021federated}, mobile computing \citep{lim2020federated}, and internet-of-things \citep{khan2021federated}.

This work considers the setting of Bayesian FL for {\em probabilistic} models. 
% involving latent variables, which has been the subject of research to date in the particular setting of fully-{\em Bayesian} models. These directly incorporate parameter uncertainty via modelling model parameters as latent variables. 
Various algorithms exist for the Bayesian FL setting, including federated \emph{stochastic variational inference} approaches such as \emph{pFedBayes} \citep{zhang2022personalized, zhang2023federated}, that estimates a \emph{global model}, which is used as a prior for each silo's \emph{local model}, somewhat regularizing each silo's inference towards a shared inference. Alternative methods include the use of Markov chain Monte Carlo (MCMC) schemes such as stochastic gradient Langevin dynamics (SGLD) \citep{plassier2021dg,el2021federated, vono2022qlsd}, distributed MCMC \citep{neiswanger2014asymptotically, wang2015parallelizing, scott2016bayes}, posterior moment matching \citep{alshedivat2021federated}, other inference methods such as expectation propagation (EP) \citep{ashman2022partitioned, guofederated2023} and surrogate likelihood \citep{jordan2018communication} approaches. 

A major limiting factor of model performance in FL settings includes \emph{heterogeneous data} among silos, and \emph{partial silo participation} -- only a proportion of silos participate at each iteration of the algorithm. Bayesian FL methods such as pFedBayes \citep{zhang2022personalized, zhang2023federated} and \emph{FedPop} \citep{kotelevskii2022fedpop} have shown promising results concerning these two issues. FedPop is the first paper to the author's knowledge that integrates ideas of mixed effect modelling \citep{gelman2006data} into the Bayesian FL area. Each silo estimates its parameters for the final layer of the Bayesian neural network, independent of the other silos. The success of these methods for heterogeneous data is due to the natural integration of the \emph{hierarchical} nature of the data into the chosen probabilistic model. 

\begin{figure}[ht]
\centering
\includegraphics[ width=0.9\textwidth]{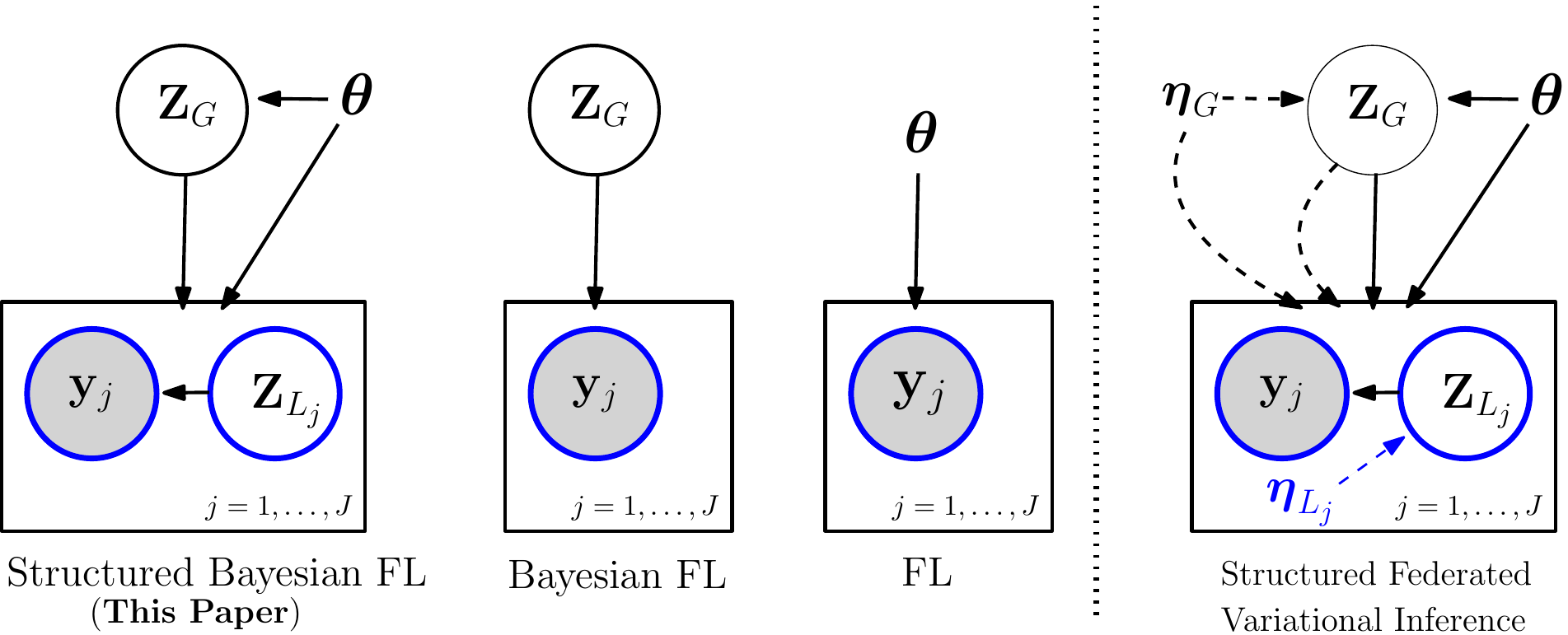}
\caption{Plate diagrams of different federated learning settings. Variables circled or written in blue are those that do not leave the silo during training. Dashed lines indicate relationships in the variational approximation.} \label{fig:plate_diag}
\end{figure}

However, the collection of existing algorithms restricts the types of appropriate models. To the authors' knowledge, no easily and widely applicable methods exist for models that do not have the structure of having {\em all} model parameters and latent variables depending on all data observations across all silos. The latter structure (illustrated in Figure \ref{fig:plate_diag}, center and rightmost) is satisfied by \emph{generalized linear models}, \emph{neural networks}, and \emph{Gaussian process} regression, yet excludes a considerable number of potential models of interest. While FedPop takes a step in this direction, it can only optimize shared parameters and constrains each silo to share the same local latent variable structure.

This work aims to extend the class of models {\em and} types of inference applicable for Bayesian FL by introducing a general approximation using variational inference techniques. The settings considered are probabilistic graphical models of the form
\begin{align} \label{eq:model_class}
\v Z_G &\sim p_{\v \theta}(\v Z_G), \\ 
\v Z_{L_j} | \v Z_G &  \stackrel{\rm ind}{\sim} p_{\v \theta}(\v Z_{L_j} | \v Z_G), \hspace{0.3cm} j=1, \ldots, J \\
\v y_j | \v  Z_G, \v Z_{L_j} & \stackrel{\rm ind}{\sim}  p_{\v \theta}(\v y_j | \v Z_{L_j}, \v Z_G), \quad j=1, \ldots, J. \label{eq:model_class_end}
\end{align}
Above, $\v Z_G$ and the collection of $\v Z_{L_j}$ are {\em latent} variables, referred to as {\em global} and {\em local} latent variables, respectively. The vector $\v \theta$ is a trainable parameter that parametrizes the probability model. We present the above with maximal generality but note that in many instances, different aspects of the model will use different subsets of $\v \theta$ or perhaps not depend on it. For example, we may have $\v \theta = (\v \theta_G^\top, \v \theta_L^\top)^\top$, where $\v \theta_G$ are prior hyperparameters for the global variables, and $\v \theta_L$ are different prior hyperparameters for the individual local variables.  In the sequel, the dimension of $\v Z_G$ and each $\v Z_{L_j}$ are denoted $n_G$ and $n_{L_j}$, respectively. Each $j$ indexes individual silo-level data $\v y_j$ (which may further decompose into individual observations $\v y_{j,1},\ldots, \v y_{j, N_j}$) and associated silo-level latent variables $\v Z_{L_j}$ (again, possibly decomposing as  $\v Z_{L_j,1},\ldots, \v Z_{L_j, N_j}$) . The objective of interest is simultaneously optimizing the marginal likelihood
\begin{equation}\label{eq:marginal_likelihood}
 p_{\v \theta}(\v y)= \int p_{\v \theta}(\v z_G)\prod_{j=1}^J p_{\v \theta}(\v y_j, \v z_{L_j} | \v z_G)d \v z_{L_j}  d \v z_G, \end{equation} 
 with respect to model parameters $\v \theta$, whilst also obtaining
posterior inferences regarding the distribution of $\v Z_G, \v Z_{L_1}, \ldots, \v Z_{L_J} | \v y$. Note that in the case that $\v \theta = \emptyset$, the task reduces to performing fully-Bayesian inference for the posterior distribution over the set of all latent variables. 
Although each $\v Z_{L_j}| \v Z_G$ must be conditionally independent, we do not require that they are identically distributed. The same holds for the distribution of each $\v y_j | \v Z_G, \v Z_{L_j}$. The above setting includes many classes of models. Some examples include hierarchical mixed models \citep{gelman2006data}, deep latent Gaussian models and variational autoencoders \citep{kingma2014autoencoding, rezende2014stochastic}, topic models in the latent Dirichlet allocation family \citep{blei2003latent, srivastava2017autoencoding}, and FedPop \citep{kotelevskii2022fedpop}, to name a few.

A further note is that due to the particular type of models under consideration, the interest is in a {\em generalized} federated learning setting where neither the data nor information regarding the posterior approximation of any local latent variables leaves its silo. The goal, unconsidered in the literature to the authors' knowledge, is to collaboratively estimate global parameters $\v\theta$ and global latent variables $\v Z_G$, yet simultaneously protect information about local latent variables $\v Z_{L_j}$ and data $\v y_j$. We impose this additional restriction as it is crucial to maintain confidentiality when estimating $\v Z_{L_j}$, as latent variables may contain sensitive information about a small proportion of observed data points. 
% This information is potentially as sensitive as the data itself. 
Figure \ref{fig:plate_diag} illustrates the distinction with previous settings. 

This work considers specifically {\em variational inference}  methods, that is, methods that replace the marginal likelihood \eqref{eq:marginal_likelihood} with a lower-bounding objective. Such methods are elegant in that they are widely applicable to different settings with minimal modifications and are more scalable than traditional inference approaches such as those involving MCMC. 

\textbf{Contribution.} The primary contributions of this work are as follows: 
(i) \emph{Structured Federated Variational Inference} (SFVI); a distributed inference algorithm that is the first approach for the structured Bayesian FL setting, which is invariant to data partitioning across silos, (ii) {\em SFVI-Avg}: A communication-efficient version of SFVI, analogous to the canonical FL algorithm {\em FedAvg} \citep{mcmahan2017communication}, and (iii)  A numerical study comparing the above two methods featuring high-dimensional classification on MNIST against, and a topic modelling example using the product latent Dirichlet allocation (ProdLDA) model.

\textbf{Paper Layout. }Section \ref{section:svi_basics} provides the requisite background surrounding stochastic-gradient VI methods. Section \ref{section:methods} introduces the SFVI algorithm, conditionally-structured variational families, and a communication-efficient SFVI-Avg algorithm. Section \ref{section:experiments} contains the numerical experiments, and Section \ref{section:discussion} concludes the paper.

{\bf Notation.} The Hadamard (elementwise) product between two vectors is denoted $\odot$. Vectors are in bold (e.g., $\v y$), and matrices are in a Romanized script (e.g., $\m X$). The notation $\v 1$ and $\v 0$ denotes a column vector of ones or zeros, respectively, of appropriate dimension. Similarly, the notation $0$ denotes a matrix of zeros where required. For a matrix $\m M$, $\text{vec}(\m M)$ denotes the vector obtained by stacking the columns of $\m M$. 

\section{Variational Inference}\label{section:svi_basics}
Variational inference is a class of methods that involve replacing the intractable marginal likelihood (model evidence) term in \eqref{eq:marginal_likelihood}, with a lower-bounding objective function that involves an approximation of the joint posterior over all latent variables, i.e., $\v Z | \v y$. Note that for the models considered herein, $\v Z = (\v Z_G^\top, \v Z_{L_1}^\top, \ldots, \v Z_{L_J}^\top)^\top$. The canonical objective for the task is the \emph{evidence lower bound} (ELBO),
\begin{align}\label{eqn:elbo}
\mathcal{L}(\v\eta, \v \theta)\coloneqq\mathbb{E}_{\v Z \sim q_{\v\eta}}[\log p_{\v \theta}(\v Z, \v y)-\log q_{\v \eta}(\v Z)].
\end{align}
Note that if $q_{\v \eta}(\v Z) \equiv p_{\v \theta}(\v Z|\v y)$, then ${\cal L}(\v \eta, \v \theta) = p_{\v \theta}(\v y)$.
In the case of fixed $\v \theta$, or where $\v \theta = \emptyset$, the above so-called {\em variational} objective is equivalent to minimizing the ${\rm KL}\big(q(\v Z)||p(\v Z | \v y)\big)$. The distribution $q_{\v \eta}$ is called the {\em variational approximation}, or the {\em variational posterior}, as it approximates the true model posterior for the latent parameters. The standard approach is to maximize the ELBO with respect to both $\v \theta$ and $\v \eta$ via stochastic optimization by estimating the gradient of the expected value in \eqref{eqn:elbo}. The gradient estimator for $\v \theta$ is straightforward, computing $\nabla_{\v \theta} \widehat{\cal L}$, where $\widehat{{\cal L}} := \log p_{\v \theta}(\v y, \v Z) - \log q_{\v \eta}(\v Z)$ for $\v Z \sim q_{\v \eta}$ is an estimator of ${\cal L}$. However, for the optimization to be practically tenable, low variance estimators of $\nabla_{\v \eta}{\cal L}$ are required. The standard strategy in modern machine learning is to perform optimization using the gradient of \eqref{eqn:elbo}. To use such an objective in a computationally feasible manner, the {\em reparameterization gradient estimator}, alternatively known as the {\em reparametrization trick}, is commonly applied for continuous distributions, which we briefly derive below. Suppose that for $\v Z \sim q_{\v \eta}$, there is a stochastic representation
$\v Z = f_{\v \eta}(\v \epsilon), \text{ } \v \epsilon \sim q_{\v \epsilon}$
for some  distribution $q_{\v \epsilon}$, such that $q_{\v\epsilon}$ does not depend on $\v \eta$. Then, writing the expectation with respect to $p_{\v \epsilon}$, and subject to mild regularity conditions permitting the exchange of the gradient and expectation operators,  a resulting (single-sample) unbiased Monte Carlo estimator of the desired gradient vector is
\begin{align}\label{eqn:stl_estimator}
\widehat{\nabla}_{\v \eta}\mathcal{L} := \deldel[f_{\v \eta}(\v \epsilon)]{\v \eta} ^\top \nabla_{\v Z}[\log p_{\v \theta}(\v Z, \v y) - \log q_{\v\eta}(\v Z)],
\end{align}
where $\v \epsilon \sim q_{\v \epsilon}$.
The estimator in \eqref{eqn:stl_estimator} is the {\em sticking the landing} (STL) gradient estimator \citep{roeder2017sticking}. Estimators of this form often have a sufficiently low variance to enable stable optimization using only a {\em single} sample per iteration. The need for a reparametrizable variational family of distributions is not particularly restrictive. Reparametrized forms exist for many distributions, including Gaussian distributions with sparsity in the precision matrix to accommodate conditional independence structures \citep{tan2020conditionally} and factor covariance structures \citep{ong2018gaussian}. Other families include normalizing flows \citep{papamakarios2021normalizing}, and implicit copula formulations 
\citep{tran2015copula, smith2021implicit}.

\section{Structured Federated Stochastic Variational Inference}\label{section:methods}
We consider variational approximations of the following generative form:
\begin{align} 
\v Z_G &\sim q_{\v \eta_G}(\v Z_G), \\ 
\v Z_{L_j} | \v Z_G &  \stackrel{\rm ind}{\sim} q_{\v \eta_{L_j}}(\v Z_{L_j} | \v Z_G), \hspace{0.3cm} j=1, \ldots, J,\label{eq:gen_variational2}
\end{align}
where $\v\eta = (\v\eta_G^\top, \v\eta_{L_1}^\top,  \ldots, \v\eta_{L_J}^\top)^\top$ are the variational parameters that parameterize the joint distribution $q_{\v \eta}$. The above choice is because the conditional independence structure of the variational approximation matches that of the target posterior distribution. Approximations with this property are referred to as {\em structured}. They are desirable as they can be more parameter efficient and lead to improved model performance \citep{hoffman2015structured, ambrogioni2021automatic}, and are often required to obtain good-quality posterior approximations \citep{tan2018gaussian, tan2020conditionally, quiroz2022gaussian}.
As we will soon demonstrate, this approximation has additional desirable properties for the model class considered in this work while satisfying the requirements of our federated learning setting discussed in the introduction. A critical insight into the structure of the model \eqref{eq:model_class}--\eqref{eq:model_class_end}, as described in Section \ref{section:introduction}, is that the required computations for the variational approximation above factorize appropriately to enable a federated inference algorithm. The appropriate calculations to demonstrate this are in the supplement. 

Algorithm \ref{alg:fedsgd} presents the  {\em Structured Federated Variational Inference} (SFVI) algorithm. We highlight that efficient computation of the vector-Jacobian product terms using automatic differentiation frameworks such as {\sf JAX} \citep{jax2018github} is straightforward.

\makeatletter
\algocf@newcmdside@kobe{ServerComputes}{%
    \KwSty{Server} \textit{computation}%
    \ifArgumentEmpty{#1}\relax{ #1}%
    \algocf@block{#2}{end}{#3}%
    % \algocf@group{#2}%
    \par
}
\newcommand\Server[2]{%
    \ServerComputes{#1}%
}
\makeatother

\textbf{\textit{Remark (SFVI is invariant to data partitioning).}} By construction, SFVI gives the {\em same} result as that obtained by performing SFVI using all the data in a {\em single} silo (or any other partitioning of the data across silos).

\begin{algorithm}[ht]
\SetAlgoLined
\KwIn{\textit{Server}: initial parameters $\v\theta$, initial variational parameters $\v \eta_G$, number of iterations $N$. \\ \quad \quad \quad \textit{Silos}:  initial variational parameters $\v\eta_{L_j}$.}
\KwOut{Parameters $\v\theta$, global variational parameters $\v\eta_G$. Each silo $j$ has access to their respective local variational parameters $\v\eta_{L_j}$.}
\For{$i=1,\ldots, N$}{
\Server{
Draw $\v \epsilon_G  \sim {\cal N}(\v{0}, \m{I}_{n_G})$\\
Send $\v\theta, \v\eta_G, \v\epsilon_G$ to each silo
}

\For{each silo $j=1, \ldots, J$ in parallel}{
Receive $\v\theta, \v\eta_G, \v\epsilon_G$ from the server\\
Draw $\v\epsilon_{L_j} \sim {\cal N}(\v 0, \m{I}_{n_{L_j}})$\\
$\v Z_G, \v Z_{L_j} \gets f_{\v \eta_G}(\v \epsilon_G), \ f_{\v \eta'_{L_j}}(\v \epsilon_{L_j})$

$\v\eta_{L_j}\gets {\sf optimizer.step}\left(\deldel[f_{\v \eta_{L_j}'}(\v \epsilon_G, \v \epsilon_{L_j})]{\v \eta_{L_j}}^\top \nabla_{\v Z_{L_j}} \log \frac{p_{\v \theta}\big(\v y_j | \v Z_G, \v Z_{L_j}\big)}{q_{\v \eta_{L_j}}}\right)$\\
$\v g_j^{\v \eta}\leftarrow  \deldel[f_{\v \eta_G}(\v \epsilon_G)]{\v \eta_G}^\top \nabla_{\v Z_G}\log \frac{p_{\v \theta}\big(\v y_j | \v Z_G, \v Z_{L_j}\big)}{q_{\v \eta_{L_j}}(\v Z_{L_j} | \v Z_G)}+ \deldel[f_{{\v \eta_{L_j}'}}(\v \epsilon_G, \v \epsilon_{L_j})]{\v \eta_G}^\top \nabla_{\v Z_{{L}_j}}\log \frac{p_{\v \theta}\big(\v y_j | \v Z_G, \v Z_{L_j}\big)}{q_{\v \eta_{L_j}}(\v Z_{L_j} | \v Z_G)}$ \\
$\v g_j^{\v \theta} \leftarrow \nabla_{\v \theta}\log p_{\v \theta}(\v y_j, \v Z_{L_{j}}| \v Z_G) $   \\
Send $\v g_j^{\v \theta}, \v g_j^{\v \eta}$ to the server
}

\Server{
Receive $\{\v g_{j}^{\v \theta}, \v g_j^{\v \eta}\}_{j=1}^J$ \\
$\v g^{\v \theta},  \v g^{\eta}  \gets  \sum_{j=1}^J \v g_j^{\v \theta}, \sum_{j=1}^J \v g_j^{\v \eta}$\\ 
$\v\theta, \v\eta_G \gets {\sf optimizer.step}\big(\v g^{\v \theta}, \v g^{\v \eta}\big)$

}

}
% note: server environment requires surrounding whitespace
\caption{SFVI}\label{alg:fedsgd}
\end{algorithm}

\subsection{Structured Gaussian Variational Family}
Structured Gaussian approximations have been shown to yield good approximations for various models (e.g., \cite{quiroz2022gaussian, tan2018gaussian}) and are readily amenable to modelling further conditionally independent structures in the individual $\v Z_{L_j}$ via sparse precision matrices.
The family of Gaussian distributions are sometimes preferred because they are easier to optimize than more sophisticated families \citep{dhaka2021challenges}. By employing ideas similar to \citet{tan2020conditionally}, who note that a fully-Gaussian approximation can be conditionally structured, we employ a joint Gaussian variational family as follows, 
\begin{align*}
    \v Z_G \sim \mathcal{N}(\v \mu_G, \Sigma_{GG}),  \text{ and } 
    \v Z_{L_j} | \v Z_G \sim \mathcal{N}\bigg(\bar{\v \mu}_j + \m C_j \v (\v Z_G- \v \mu_G), \Sigma_{L_j L_j}\bigg),
\end{align*} 
for $j=1,\ldots, J$. It is straightforward to show that the above yields that $\v Z = (\v Z_G^\top, \v Z_{L_j}^\top)^\top$ is jointly  Gaussian with $\bb C{\rm ov}(\v Z_G, \v Z_{L_j})=  \Sigma_{GG}C_j$. The reparametrized generative form of the above used in the experiments takes $(\v \epsilon_G^\top, \v \epsilon_{L_1}^\top, \ldots, \v \epsilon_{L_J}^\top) \sim \mathcal{N}(\v 0, \m I)$, and then sets 
$\v Z_G= \v \mu_G + \v \sigma_G \odot \m L_{G}\v \epsilon_G$, and $\v Z_{L_j} = \bar{\v \mu}_j + \m C_j \v (\v Z_G- \v \mu_G) + \v \sigma_j \odot \m L_j \v \epsilon_{L_j}$, for $j=1,\ldots,J$, where $\m L_G \in \bb R^{n_G \times n_G}$ and each $\m L_j \in \bb R^{n_{L_j}\times n_{L_j}}$ are lower-unitriangular. 

\subsection{SFVI-Avg}\label{section:federated_averaging}
SFVI focuses primarily on enabling aspects of
probabilistic FL by extending the class of models for
which federated analysis is possible. Here, we consider a {\em communication-efficient} variant of SFVI for the structured model setting. As mentioned, one can view SFVI as an analogue of the baseline {\em FedSGD} FL algorithm considered in \citet{mcmahan2017communication}, who proceed to propose an alternative algorithm called {\em federated averaging} (FedAvg). The latter involves training a model on individual silos using only data available to the individual silos whilst periodically sharing the model parameters with a central server that aggregates, averages and sends back updated model parameters to the silos to repeat the procedure. \citet{mcmahan2017communication} show that their approach works well for neural network training, which applies only to the standard FL model setting (Figure \ref{fig:plate_diag}, second from the right). 

Here, an algorithm, {\em SFVI-Avg}, that considers such ideas in the structured Bayesian FL setting is presented. We begin by noting that for the global parameters $\v \theta$, one can average as usual as in {\em FedAvg}. However, we consider three additional aspects required for the algorithm to be suitable for variational parameters:

\begin{enumerate}
    \item {\bf Only quantities regarding the posterior variational approximation of $\v Z_G$ need to be averaged}. The local latent variables are only of interest to their silos; their posterior distributions depend only on the data within their silo and the global values $\v Z_G$. Thus if models are fit to an individual silo's data, only the {\em global} latent variable values need to be averaged. 
\item {\bf For averaging to make sense, the model requires additional structure.} While SFVI permits a very general model structure, for any averaging approximation of {\em distributions} to be accurate concerning the global latent variables $\v Z_G$, all models must use the same likelihood for all $N$ individual observations contained within $\v y$. Sacrificing some generality to capture this effect, we shall assume the simple setting that collections of observations among different silos share a similar latent variable structure and the same likelihood. Specifically, the model takes the form, 
\begin{align*} 
\v Z_G &\sim p_{\v \theta}(\v Z_G), \\ 
\v Z_{L_{j,k}} | \v Z_G &  \stackrel{\rm iid}{\sim} p_{\v \theta}(\v Z_{L_{j,k}} | \v Z_G), \hspace{0.3cm} j=1, \ldots, J, \quad k=1,\ldots, N_j \\
\v y_{j,k} | \v  Z_G, \v Z_{L_{j,k}} & \stackrel{\rm iid}{\sim}  p_{\v \theta}(\v y_{j,k}| \v Z_{L_{j,k}}, \v Z_G), \quad j=1, \ldots, J, \quad k=1,\ldots, N_j. \label{eq:model_class_end}
\end{align*}
The justification is that in such settings, by multiplying the log-prior terms for $\v Z_{{L}_j}|\v Z_G$ and the log-likelihood by a factor of $N/N_j$ where $N$ is the total number of observations across all silos, and $N_j$ are those in the silo currently fitting the model, an approximation to the overall scale of the joint log-density function across all silos is, 
 \[\log p_{\v \theta}(\v Z_{L}, \v y| \v Z_G) \approx \frac{N}{N_j}\sum_{k=1}^{N_j} \nabla_{\v \theta}\log p_{\v \theta}(\v Z_{L_{j,k}}, \v y_{j ,k}| \v Z_G).\]
\item \textbf{The {\em type} of averaging matters.} Simply taking parameter averages of the variational parameters will yield strikingly different results depending on the parametrization of the variational family. We thus consider the more principled approach of averaging in {\em distributional} space instead of {\em parameter} space. Wasserstein barycenters have been used for posterior averaging in distributed MCMC algorithms \citep{srivastava2018scalable, ou2021scalable}. We use Wasserstein barycenters to average the variational approximation of the global latent variables across silos. Our algorithm is the first to consider such an averaging scheme in the federated learning setting. For a collection of measures $\{\pi_j\}_{j=1}^J$, and ${\cal W}(\nu, \pi)$ denoting the Wasserstein metric \citep{panaretos2019statistical}, the  {\em Wasserstein barycenter} is defined to be 
$\nu^\star = \arg \min_{\nu} \sum_k  \mathcal{W}(\nu, \pi_k)^2$. 
In the case where the elements of $\{\pi_j\}_{j=1}^J$ are all multivariate Gaussian, the barycenter is unique and is itself Gaussian \citep[Theorem 4]{mallasto2017learning}, with mean vector 
$\v \mu_\star = J^{-1} \sum_j  \v \mu_j$, and covariance matrix that is the unique solution to the root-finding problem
$\Sigma_\star = J^{-1} \sum_j(\Sigma_\star^{1/2} \Sigma_j \Sigma_\star^{1/2})^{1/2}$. The latter can be computed numerically via fixed-point iteration \citep{alvarez2016fixed}); an implementation is available in the  \verb|ott| package \citep{cuturi2022optimal}. Moreover, in the case in which each $\Sigma_j$ is diagonal, an analytical solution is available, given by the diagonal matrix 
$\m \Sigma_{\star} = \left(J^{-1}\sum_{j} \m \Sigma_j^{1/2} \right)^2$. 
\end{enumerate}

Taking the above points into account, Algorithm \ref{alg:fedavg} presents {\em structured federated variational inference with averaging} (SFVI-Avg), indexed by a strictly-positive integer $m$ that determines the number of {\em local} iterations at each step before averaging. 

\begin{algorithm}[ht]
\SetAlgoLined
\KwIn{\textit{Server:} initial parameters $\v\theta$, initial global variational parameters $\v \eta_G$.\\ \textit{Silos:} initial local variational parameters $\v\eta_{L_j}$, total observations $N$, and number of observations for each silo $N_j$, number of rounds $R$.}
\KwOut{Parameters $\v\theta$, variational parameters $\v\eta_G$, and $\v\eta_{L_j}$ (the latter held by each silo $j$).}
\For{$i=1,\ldots, R$}{
\Server{
Send $\v\theta, \v\eta_G$ to each silo
}\\

\For{each silo $j=1, \ldots, J$ in parallel}{
Receive $\v\theta, \v\eta_G$ from the server\\
$\v\theta^{(j)}, \v\eta_G^{(j)}\gets \v\theta, \v\eta_G$\\
Update $\left(\v \theta^{(j)},\v \eta_G^{(j)}, \v \eta_{L_j}\right)$ via $m$ optimization steps of {\em local} stochastic-gradient VI using \eqref{eqn:stl_estimator} with 
\[\begin{split}\nabla_{\v \theta}\log p_{\v \theta}(\v Z) := \nabla_{\v \theta}\log p_{\v \theta}(\v Z_G) + \frac{N}{N_j}\nabla_{\v \theta}\log p_{\v \theta}(\v Z_{L_j}, \v y_j | \v Z_G)\end{split}\] 
 \\
Send $\v\theta^{(j)}, \v\eta_G^{(j)}$ to the server
}

\Server{Receive $\left \{\v\theta^{(j)}, \v\eta_G^{(j)}\right \}_{j=1}^J$\\
$\v\theta  \gets J^{-1}\sum_{j=1}^J \v\theta^{(j)}$ \\
 $\v\eta_G \gets$  Variational parameters corresponding to the Wasserstein barycenter of distributions induced by the parameters $\v \eta_G^{(j)}$ for $j=1,\ldots,J$.}

}
 
\caption{SFVI-Avg($m$)}\label{alg:fedavg}
\end{algorithm}

\textbf{Remark (Amortized Inference)}. Both SFVI and SFVI-Avg readily extend to the amortized inference setting \citep{kingma2014autoencoding, rezende2015variational}. Here, the variational parameters for the local latent variables are not trained directly but are instead defined via some neural network $f_{\v \phi}$ (parameterized by $\v \phi$) called an \textit{inference network}. The inference network parameters $\v \phi$ are trained instead of directly training the variational parameters $\v \eta_{L_j}$ (in our notation, $\v \phi \in \v \theta$). 
The required modifications to the algorithms are straightforward; one simply sets $\v \eta_{{L}_{j,k}} = f_{\v \phi}(\v y_{j,k}, \v Z_G)$
and hence $\log q(\v Z_{L_j}|\v Z_G) = \sum_{k=1}^{N_j} \log q(\v Z_{L_j, k}; f_{\v \phi}(\v y_{j,k}))$. 

\section{Numerical Experiments}\label{section:experiments}
This section presents a series of numerical examples. Their primary aim is to demonstrate the simplicity and flexibility with which the proposed algorithms can fit complex models that could not be used in the FL settings until now.  A parallel aim is to assess the performance of the proposed {\em SFVI-Avg} algorithm compared to {\em SFVI}. 

It is worth highlighting that our algorithm(s) can potentially aid the development of complex hierarchical models that can be used in the heterogeneous FL setting, for which inference algorithms have hitherto been unavailable. Such models and inference approaches are considered in the first example. However, while the considered models perform reasonably well and, in some cases, better than two state-of-the-art methods, it is worth highlighting that we view the primary contribution of this work as providing algorithms that can be used to fit a general class of probabilistic models, unable to be fit in the FL setting up to this point. Determining the optimal hierarchical deep learning model for which to obtain the best results for a federated learning task is beyond the scope of this paper and warrants further investigation. 

All experiments use the {\sf adam} optimizer \citep{kingma2015adam} as implemented in the {\sf optax} package \citep{deepmind2020jax}. Note that the experiments implicitly compare to a variational approximation using the full data on a single silo, as this is what is given by the SFVI algorithm (see earlier remark). 

Additional details about the experiments, as well as additional experiments, can be found in the supplement.  
 
\subsection{Hierarchical Bayesian Neural Network for Heterogeneous Data}

This example uses the MNIST dataset of handwritten digits \citep{lecun1998mnist} in a setting with heterogeneous data partitions. We partition the training and test sets into either 10 or 50 silos, each containing an equal number of observations. However, 90\% of each silo's observations correspond to a single digit (label), with the remaining nine digits being represented approximately uniformly among the remaining 10\% of observations. For $i=1,\ldots,784,  k=1,\ldots,64$, and $j=1,\ldots,10$, the model is
$\mu_{ik}  \sim \mathcal{N}(0,1)$, $\sigma \sim  \mathcal{N}_+(0,1)$, $\epsilon_{ik}^{(j)} \sim_{\rm ind} \mathcal{N}(0,1)$, $W^{(1,j)}_{ik}$ = $\mu_{ik} + \sigma \epsilon_{ik}^{(j)}$, and $W^{(2, j)}_{ik} \sim_{\rm iid} \mathcal{N}(0 ,1)$. Then, the {\em personalized} neural network model for silo $j$ is 
\[f_j(\v x) = {\rm softmax}\left(\m W^{(2,j)}{\rm ReLU}\left(\m W^{(1,j)}\v x\right)\right).\]
In the above setting,  $\boldsymbol{Z}_G =((\mu_{ik}), ({\sigma_{ik}})) $, 
$\boldsymbol{Z}_{L_j}= \left(\m W^{(1,j)}, \left(\epsilon_{ik}^{(j)}\right)\right)$ for $j=1,\ldots,J$, and $\boldsymbol{\theta}=\emptyset$. 

We call this {\em hierarchical BNN}, as the first layer has a hierarchical structure by sharing a Gaussian prior for the weights of the first layer across silos using a {\em non-centered parameterization}. Such modelling approaches are well-known to be effective in modelling data heterogeneity. 
For further details, see, for example, \cite{gelman2006data}.

Using a BNN with silo-personalized latent variables for heterogeneous data in the FL setting was introduced in the so-called FedPop model \citep{kotelevskii2022fedpop}. However, limitations of the associated inference algorithm proposed, FedSOUL (again, in \citep{kotelevskii2022fedpop}) are that the algorithm provides maximum a-posteriori estimates for the global latent variables $\boldsymbol{Z}_G$ (as opposed to fully-Bayesian inferences) and that the neural net architecture must be the same across silos.

The proposed inference algorithms, SFVI and SFVI-Avg, do not have these limitations and can perform inference on the Hierarchical BNN proposed and a fully-Bayesian version of the FedPop model. We also compare with pFedBayes \citep{zhang2022personalized} on this example.   

Table \ref{hierBNN_results} shows the comparative performance between methods over five (5) independent runs, each involving different partitions of the data among silos created as described at the beginning of this subsection. Of note is that SFVI o performs all FL algorithms. The FedSOUL algorithm sings the FedPop performs the best concerning test predictive accuracy out of the FL algorithms. However, the standard error of the predictive test accuracy of the FedSOUL algorithm across each of the silo's test predictive accuracy is much greater than either of the models that were fit using SFVI-Avg. 

\begin{table}[]\label{hierBNN_results}
\centering

\caption{Test accuracy results for the MNIST example with severe data heterogeneity across silos over five runs. FedSOUL and pFedBayes were run with standard parameters.}
\label{tbl:BNNresults}
\begin{tabular}{c|c|c|c|c}
& & $J=10$ & $J=50$
\\
\textbf{Model   }         & \textbf{Inference} & \textbf{Acc.} \% (std) & \textbf{Acc.} \% (std) & \textbf{Rounds} \\ \hline 
Hierarchical BNN & SFVI -Avg      &     96.6 (0.68)               & \textcolor{red}{\textbf{93.6}} (1.68) & 20      \\
 % & & \textbf{97.1} (0.80) & & 50 \\  
         & SFVI             & \textbf{97.5} (0.50)                &  96.0 (1.03)   &   $10^4$ \\ \hline 

Fully-Bayesian FedPop  
& SFVI-Avg         &          97.0 (0.62)     & 94.2 (1.33)     & 20 \\ 
& SFVI             &       \textbf{97.5} (0.46)             & 96.2 (1.05) &     $10^4$ \\ \hline 
         
FedPop           & FedSOUL            &          97.3 (1.41)          &  95.4 (8.57)  &  \textbf{20}     \\
\citep{kotelevskii2022fedpop}    &  & & & \\ \hline  
pFedBayes        &         pFedBayes          &         {\bf \textcolor{red}{94.1}} (2.20)           &  \textbf{97.1} (1.18) &   10    \\ \citep{zhang2022personalized}
                 &                  &                    &        \\  
      % Baseline BNN           &       SFVI-Avg           &    90.0 (3.59)               &        \\   & SFVI & 96.0 (0.04)  &  
\end{tabular}
\end{table}

\subsection{Product Latent Dirichlet Allocation}
The second experiment considers a topic modelling example. ProdLDA \citep{srivastava2017autoencoding} is a slight modification of the original latent Dirichlet allocation model of \citet{blei2003latent}, for which \citet{srivastava2017autoencoding} demonstrates that variational inference methods on ProdLDA outperform classic LDA, even when collapsed-Gibbs MCMC sampling methods are used for the latter. The model is 
$$\v T_j \sim_{\rm iid} {\rm Dirichlet}(\beta \mathbf{1}_{n_{\rm vocab}}), \quad j=1,\ldots,n_{\rm topics}, $$
$$\v W_k \sim_{\rm iid} \mathcal{N}(\alpha {\bf 1}_{n_{\rm topics}},1) , \quad  k = 1,\ldots, n_{\rm docs},$$
\[\v c_{k}| \m T, \v W_k \sim_{\rm ind} {\rm Multinom}(l_k, {\rm softmax}(\m T \v W_k )), \quad k = 1,\ldots, n_{\rm docs}\]
where  $T := (\v T_1,\ldots, \v T_{n_{\rm topics}}) \in \bb R^{n_{\rm vocab} \times n_{\rm topics}}$, and $l_k$ denotes the length of (number of word tokens in) document $k$. Above, $\v \theta = (\alpha, \beta)$, $\v Z_G = (\v T_1^\top,\ldots, \v T_{n_{\rm topics}}^\top)^\top$, and $\v Z_{L} = (\v W_1^\top,\ldots, \v W_{n_{\rm docs}}^\top)^\top$. Note that the total dimension is $n_{\rm vocab} \times n_{\rm topics}$, and that $\v Z_{L}$ is $n_{\rm topics} \times n_{\rm docs}$ dimensional. 

The example uses the 20Newsgroups dataset, which contains $n_{\rm docs}=18884$ messages obtained from 20 different online newsgroups. The data are preprocessed (stop words are removed, and tokens are lemmatized using the $\textsf{gensim}$ package \cite{rehurek_lrec}) and randomly split equally into three ($3$) separate imaginary silos. The vocabulary of words is chosen to be the most common $2000$ words in the corpus, i.e., $n_{\rm vocab} = 2000$, and the number of topics $n_{\rm topics} = 21$. Given the high-dimensionality of the posterior distribution, the overall approximating family is chosen to be a multivariate Gaussian with {\em diagonal} covariance matrix. 

\begin{figure}[ht]\label{coherence}
    \centering
    \subfigure[Individual topic coherence values for the ProdLDA model trained using different approaches (higher values are better).]{
        \includegraphics[width = .45\textwidth]{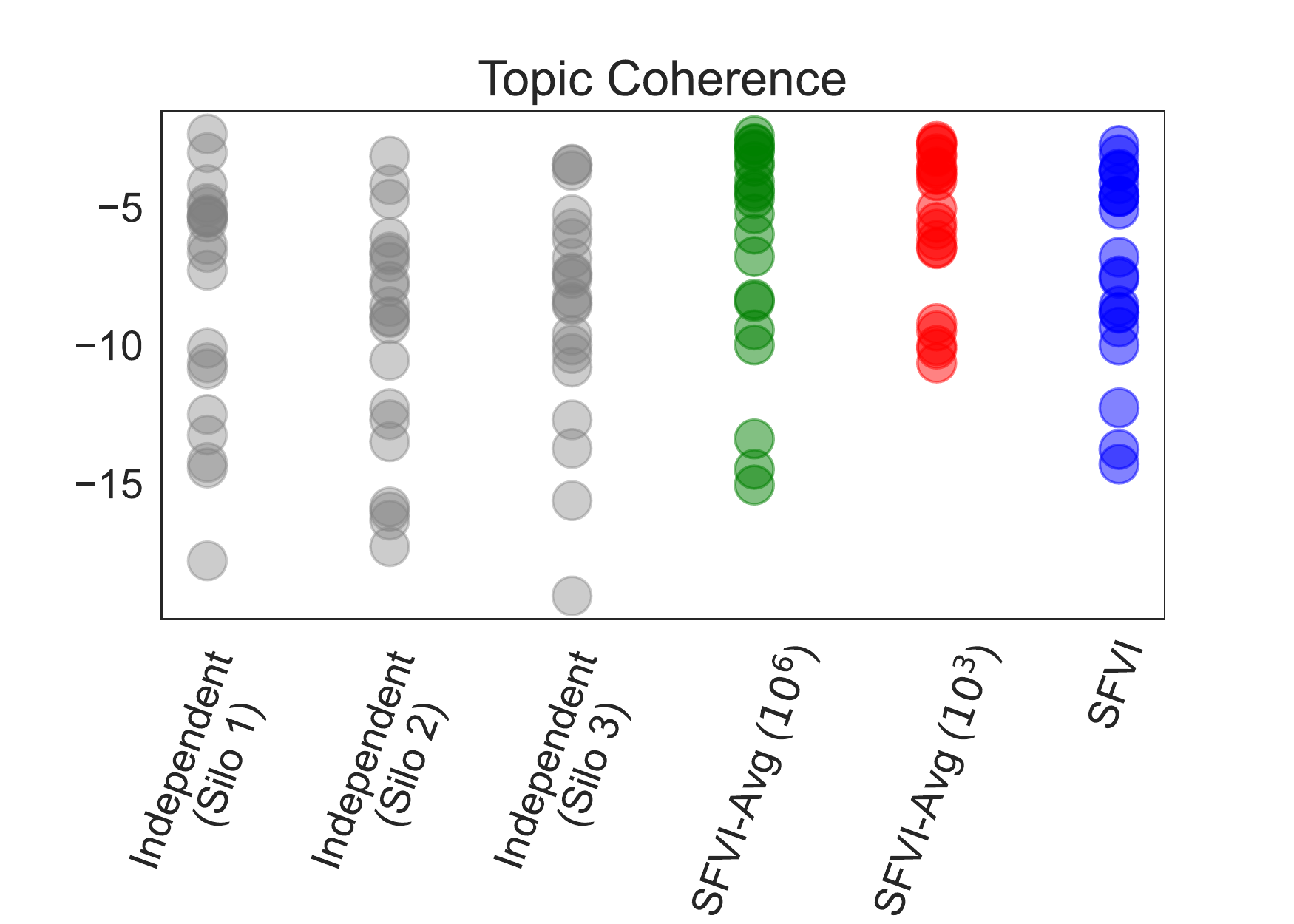}
        \label{fig:subfig1}
    }
    \hspace{5mm}  % Add horizontal space between the figures
    \subfigure[ELBO values during training for the ProdLDA example on the {\sf 20NewsGroups} dataset. The number of iterations for SFVI reported in the plot is $10^3$ times that reported on the $x$-axis.]{
    % Add vertical space between the caption and the figure
        \includegraphics[width=.45\textwidth, trim= 30 -52 30 30]{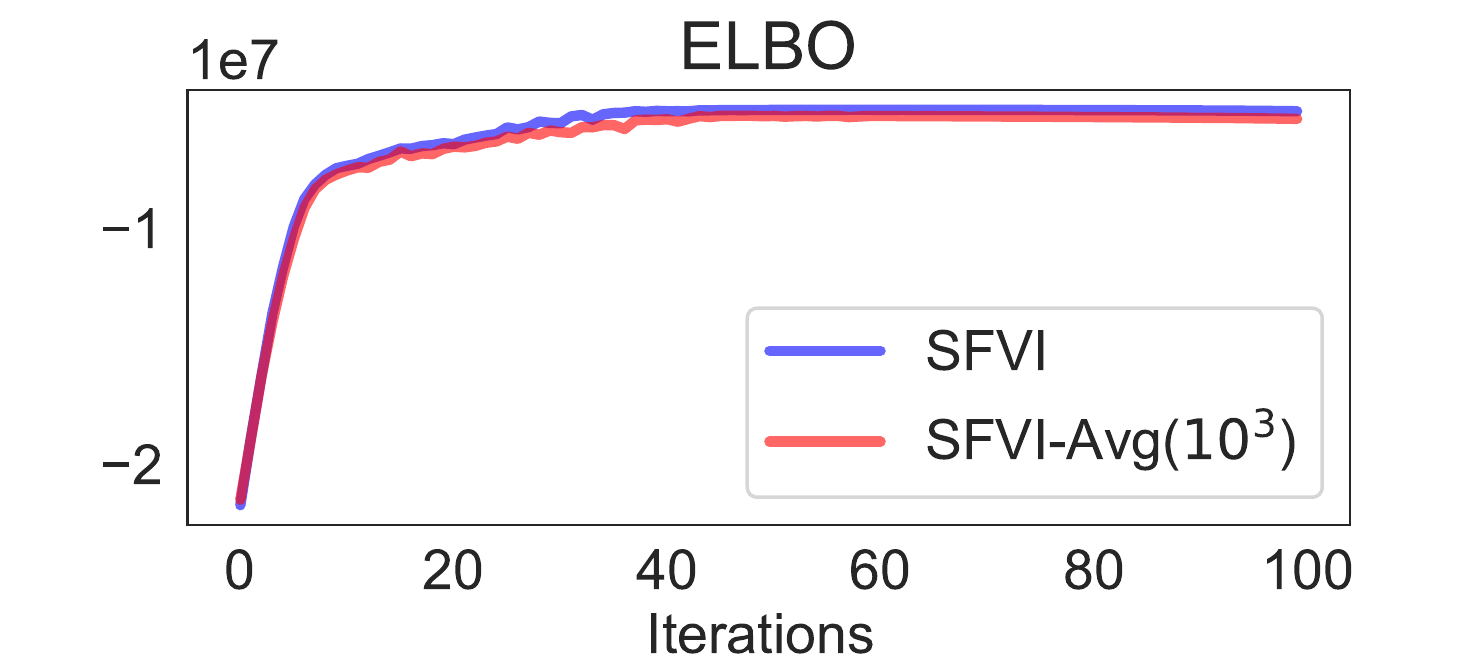}
        \label{fig:subfig2}
    }
    \caption{ProdLDA topic coherence and ELBO values.}
    \label{fig:main}
\end{figure}

To benchmark model quality across methods, we plot the {\em (UMass) Coherence Score} \citep{mimno2011optimizing} across topics. Figure \ref{coherence} (a) plots the individual topic coherence values for different training approaches. The results demonstrate that our proposed methods outperform fitting the model to the data in each silo. The SFVI-Avg algorithm with $10^2$ communication rounds and $10^3$ local steps outperforms SFVI in terms of topic coherence despite the latter attaining a higher ELBO (shown in Figure \ref{coherence} (b)).  

\section{Discussion}\label{section:discussion}
The paper has extended the models for which federated inference is possible by proposing a simple and general solution with a flexible yet judicious choice for the approximating family. An initial step towards communication-efficient inference in the above setting was also explored, with promising results. To the author's knowledge, the approaches considered herein are the most general and easily applicable methods for structured probabilistic models. 

The advent of structured probabilistic models in the FL context presents a significant advancement with extensive potential applications. With the inherent complexity and heterogeneity in real-world data, structured models allow for incorporating diverse and intricate relationships among variables, thus capturing the underlying data structure more accurately. The hierarchical representation in these models aligns perfectly with the FL setting, where data is naturally partitioned across various silos and offer a robust way to address the significant problem of data heterogeneity in FL.

There are many possible extensions to the work herein in several directions. Firstly, while the paper focused on the ELBO objective, the canonical variational objective, the ideas herein easily translate to other objectives such as the importance-weighted objective \citep{burda2016importance} with doubly-reparametrized gradient estimators \citep{tan2020conditionally} or locally-enhanced bounds \citep{geffner2022variational}. An interesting direction is obtaining formal mathematical guarantees surrounding the privacy of information inclusive of either data or local latent variables via the {\em differential privacy} framework. Such an avenue may include applying DP-SGD \citep{abadi2016deep} and extending ideas from previous differentially-private variational inference methods \citep{jalko2016differentially}. A theoretical investigation of the properties of SFVI-Avg, for example, in terms of convergence, is also of interest, as are other refinements. For example, improved averaging schemes based on concepts of density-ratio estimation \citep{sugiyama2012density} may prove promising. 
\bibliographystyle{plainnat}
\bibliography{references}
%\putbib[references]
%\end{bibunit}
\part*{Supplementary Material}

\setcounter{section}{0}
\setcounter{figure}{0}
\setcounter{table}{0}
\setcounter{equation}{0}
\renewcommand{\thesection}{S\arabic{section}}
\renewcommand{\thefigure}{S\arabic{figure}}
\renewcommand{\thetable}{S\arabic{table}}
\renewcommand{\theequation}{S\arabic{equation}}

\section{Derivation of SFVI Updates}
First, note that as a consequence of the structure
\begin{align} \label{eq:gen_variational}
\v Z_G &\sim q_{\v \eta_G}(\v Z_G), \\ 
\v Z_{L_j} | \v Z_G &  \stackrel{\rm ind}{\sim} q_{\v \eta_{L_j}}(\v Z_{L_j} | \v Z_G), \hspace{0.3cm} j=1, \ldots, J,
\end{align} the reparametrized variables necessarily decompose where
$\v \epsilon = \left(\v \epsilon_G^\top, \v \epsilon_{L_1}^\top, \ldots, \v \epsilon_{L_J}^\top \right)^\top$ 
and 
\[
\v Z_{\v \eta_G}\coloneqq f_{G}(\v\epsilon_G),\quad \v Z_{L_j} \coloneqq f_{\v \eta_{L_j}'}(\v \epsilon_{G}, \v \epsilon_{L_j}),\,\, j=1, \ldots, J,
\] 
where $\v \eta_{L_j}' :=  (\v \eta_G^\top, \v \eta_{L_j}^\top)$, and thus $f_{\v \eta}(\v \epsilon)$ is equal to 
\[\left( f_{\v \eta_G}(\v \epsilon_G)^\top, f_{\v \eta_{L_1}'}(\v \epsilon_G, \v \epsilon_{L_1})^\top, \ldots, f_{\v\eta_{L_J}'}(\v \epsilon_G, \v \epsilon_{L_J})^\top \right)^\top,\]
with the transposed Jacobian matrix having a block upper-triangular structure of the form
\[
\begin{pmatrix}
\deldel[f_{\v \eta_G}(\v \epsilon_G)]{\v \eta_G} ^\top \quad \ \     & \deldel[f_{\v \eta_{L_1}'}(\v\epsilon_G, \v\epsilon_{L_1})]{\v \eta_G} ^\top  & \cdots & \deldel[f_{\v \eta_{L_J}'}(\v \epsilon_G, \v \epsilon_{L_J})]{\v \eta_G}^\top \\
\m 0 &\deldel[f_{\v \eta_{L_1}'}(\v\epsilon_G, \v\epsilon_{L_1})]{\v \eta_{L_1}}^\top &  \vdots & 0    \\
\vdots & 0 &  \ddots   & \m 0\\ 
\m 0 & 0 &  \cdots   & \deldel[f_{\v \eta_{L_J}'}(\v \epsilon_G, \v \epsilon_{L_J})]{\v \eta_{L_J}}^\top
\end{pmatrix}.
\]

Due to the above sparsity structure, the STL gradient estimator  \citep{roeder2017sticking} 
\begin{align}\label{eqn:stl_estimator2}
\widehat{\nabla}_{\v\eta}\mathcal{L} := \deldel[f_{\v\eta}(\v\epsilon)]{\v\eta}^\top\nabla_{\v Z}[\log p_{\v\theta}(\v Z, \v y) - \log q_{\v\eta}(\v Z)], 
\end{align} decomposes as
\[\widehat{\nabla}_{\v \eta}{\cal L} = \left(\big(\widehat{\nabla}_{\v\eta_G} {\cal L}\big)^\top, 
(\widehat{\nabla}_{\v\eta_{L_1}} {\cal L})^\top,\ldots, 
\big(\widehat{\nabla}_{\v\eta_{L_J}} {\cal L}\big)^\top
  \right)^\top,\]
where the terms on the left hand side above are defined analogously to \eqref{eqn:stl_estimator2}. Next, put 
\[ 
\widehat{\mathcal{L}}_0 := \log \frac{p_{\v \theta}\big(\v Z_G)}{q_{\v \eta_G}(\v Z_G)}, \text{and }  
\widehat{\mathcal{L}}_j := \log \frac{p_{\v \theta}\big(\v y_j, \v Z_{L_j} | \v Z_G)}{q_{\v \eta_{L_j}}(\v Z_{L_j} | \v Z_G)},
\]
for $j=1,\ldots, J$, noting that $\widehat{{\cal L}} = \sum_{j=0}^J \widehat{\cal L}_j$. Following some algebra and removal of terms that vanish for certain gradients, one obtains that
\begin{equation}\label{eq:identity1}
\widehat{\nabla}_{\v\eta_G} {\cal L} =  \deldel[f_{\v \eta_G}(\v \epsilon_G)]{\v \eta_G}^\top \nabla_{\v  Z_G}\widehat{\cal L}_0 + \sum_{j=1}^J \v g_j^{\v \eta},
\end{equation}
where
\begin{equation}\begin{split}
\v g_j^{\v \eta} :=& \deldel[f_{\v \eta_G}(\v \epsilon_G)]{\v \eta_G}^\top \nabla_{\v Z_G}\widehat{\mathcal{L}}_j + \deldel[f_{{\v \eta_{L_j}'}}(\v \epsilon_G, \v \epsilon_{L_J})]{\v \eta_G}^\top \nabla_{\v Z_{{L}_j}}\widehat{{\cal L}}_j.
\label{eq:g_j}
\end{split}\end{equation}
Finally, a straightforward computation yields for $j=1,\ldots,J$, 
\begin{equation}\begin{split}
 \widehat{\nabla}_{\v\eta_{L_j}}^{\rm STL} {\cal L} = & \deldel[f_{\v \eta_{L_j}'}(\v \epsilon_G, \v \epsilon_{L_j})]{\v \eta_{L_j}}^\top \nabla_{\v Z_{L_j}} \widehat{\mathcal{L}}_j.
\label{fedsgd:local_grad}
\end{split} \end{equation}  
From the above, note that the only required terms to update $\v \eta_{L_j}$ are $\v \epsilon_G$ and $\v \eta_G$. Further, the only terms required to update $\v \eta_G$ are the $\v g_j$ as defined in \eqref{fedsgd:local_grad} which do {\em not} require knowledge of the $\v Z_{L_j}$ simulated from the variational approximation, nor the parameters of the distribution that generated them. Finally, the gradient updates with respect to $\v \theta$ simplify as 
\begin{equation}\begin{split}\nabla_{\v \theta}\widehat{{\cal L}} = \log p_{\v \theta}(\v Z_G) +  \sum_{j=1}^J \v g_j^{\v \theta}, \end{split}
\end{equation}
where
 \begin{equation}\v g_j^{\v \theta} := \nabla_{\v \theta}\log p_{\v \theta}(\v y_j, \v Z_{L_{j}}| \v Z_G). \label{eq: g_j_theta}\end{equation}
Thus, silos need only share $\v g_j^{\v \theta}$ with the server for the required updates of $\v \theta$ to be performed. The algorithm follows directly from the simplifications above.

\section{Details for Experiments in the Paper}
\subsection{Hierarchical Bayesian Neural Networks}
For the results of SFVI and SFVI-Avg, a diagonal Gaussian variational approximation is used in all experiments. The results of pFedBayes and FedPop use the default hyperparameters used for MNIST data in their respective papers. 

% \begin{itemize}
% \item \textbf{SFVI-Avg:} mean-field variational Bayes of the BNN with local last layers using the SFVI-Avg algorithm with $20$ global rounds and $100$ local updates per round. \\[-1.85em]
% % \item \textbf{SFVI-Avg (p):} equivalent to \textbf{SFVI-Avg}, but each silo has a 50\% chance of participating in each global round.\\[-1.85em]
% % \item \textbf{Fed-Avg:} mean-field variational inference using the standard BNN with $20$ global rounds and $100$ local updates per round.\\[-1.85em] 
% \item \textbf{pFedBayes:} Hyperparameters set to $\rho=-3$, $\zeta=10$, and $\eta=0.001$, with $10$ global rounds and $20$ local updates per round.
% \item \textbf{FedPop:} Default hyperparameters in the original implementation.
% \end{itemize}

\section{Additional Experiments}
\subsection{Bayesian Logistic Mixed Model}
This example considers a fully-Bayesian analysis of a logistic mixed model. Here, we consider the six cities' dataset \citep{fitzmaurice1993likelihood}, extracted from a longitudinal study of $537$ children, to assess the health effects of air pollution. A binary response $y_{ij}$, representative of the presence of wheezing, was recorded yearly from the age of 7 to 10, where $i$ indexes the child and $j$ indexes the four observations for each child. There are two covariates: a binary indicator for the smoking status of the mother, $x_i^{\rm smoke}$, and the current age of the child, centred at age nine, $x_{ij}^{\rm age}$. The model is,
\[\begin{split}
y_{ij}| \v \beta, Z_{i} &\sim_{\rm ind} {\rm Bern}(p_{ij}), \quad i=1,\ldots537, \  j=1,\ldots,4, \\
p_{ij} &= {\rm logit}^{-1}\big( \beta_0 + \beta_{1}x_i^{\rm smoke}  + \beta_{2}x_{ij}^{\rm age} + \beta_{3}x_i^{\rm smoke}\cdot x_{ij}^{\rm age} + b_i \big),\\
\beta_k &\sim_{\rm iid} \mathcal{N}(0, 10^2), \quad k=0,\ldots, 3,  \\ 
 \omega &\sim  {\cal N}(0, 10^2), \\ 
 b_{i} | \omega &\sim_{\rm iid} {\cal N}(0,  \exp(-2\omega)), \quad i=1,\ldots,537.
\end{split}
\]
Above, each $b_i$ is a {\em random effect} term, taking the form of a random {\em intercept}. In this example, we have that  $\v Z_G = (\v \beta^\top, \omega)^\top$ , $\v Z_L = \v b$, and $\v \theta = \emptyset$. To further capture the target posterior's structure, we set $\m L_j \equiv \m I$ as each $\v b_i$ is conditionally independent a posteriori given $\v Z_G$ and the observed data.  
We randomly split the data into two silos containing $300$ and $237$ children. Despite the apparent simplicity of this model, it is known to be challenging even for \citet{tan2020conditionally}, and mixed models, in general, require structured approximations to obtain reasonable posterior marginal approximations in a fully-Bayesian setting (e.g., \cite{tan2018gaussian}). Moreover, an uneven data split tends to yield vastly different inferences for the global variables when silos fit data independently. These points, combined with the small-data setting, make such an example a challenge for {\em SFVI-FedAvg}.

\begin{figure}[]
    \centering
    \includegraphics[width = .8\textwidth]{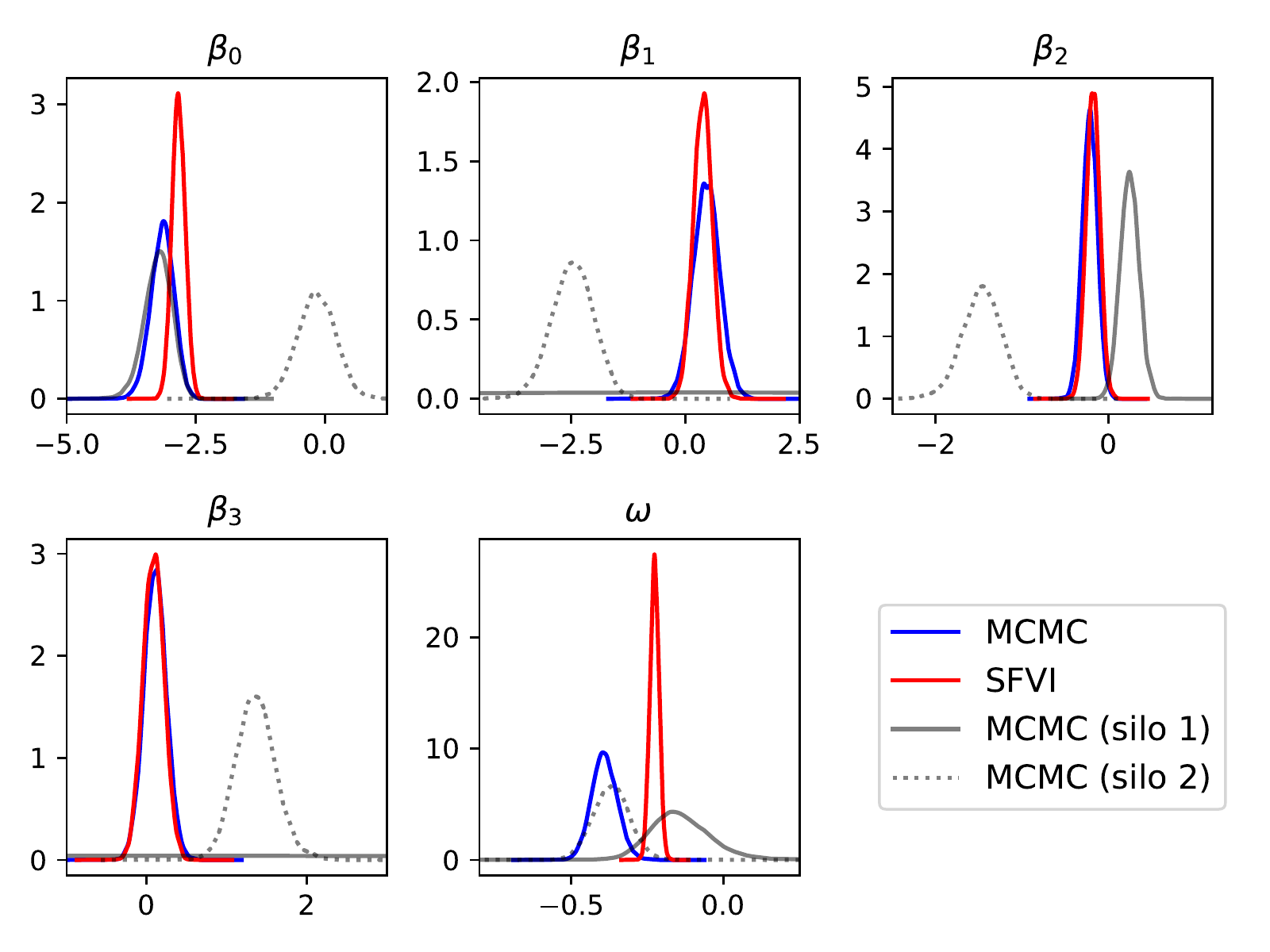}
    \caption{Marginal posterior parameters for the Bayesian GLMM example.}
    \label{fig:GLMM_marginals}
\end{figure}

We compare SFVI to MCMC (using the No-U-Turn sampler \citep{hoffman2014no} as implemented in {\sf NumPyro} \citep{phan2019composable}) fit on all data, and the independent silos. Figure \ref{fig:GLMM_marginals} shows that SFVI  captures the marginal posteriors of the regression parameters $\v\beta$ accurately, irrespective of minimal overlap in marginal densities of the independent silos. Such results were consistent across runs with different seeds. Regrettably, {\rm SFVI-Avg} failed to produce reasonable results for this example, illustrating that despite the very promising results in 
the other examples, it is not a panacea for communication-efficient federated inference in all settings. 

\subsection{Empirically-Bayesian Multinomial Regression}
\begin{figure}[ht]
    \centering
    \includegraphics[width = 0.6\textwidth]{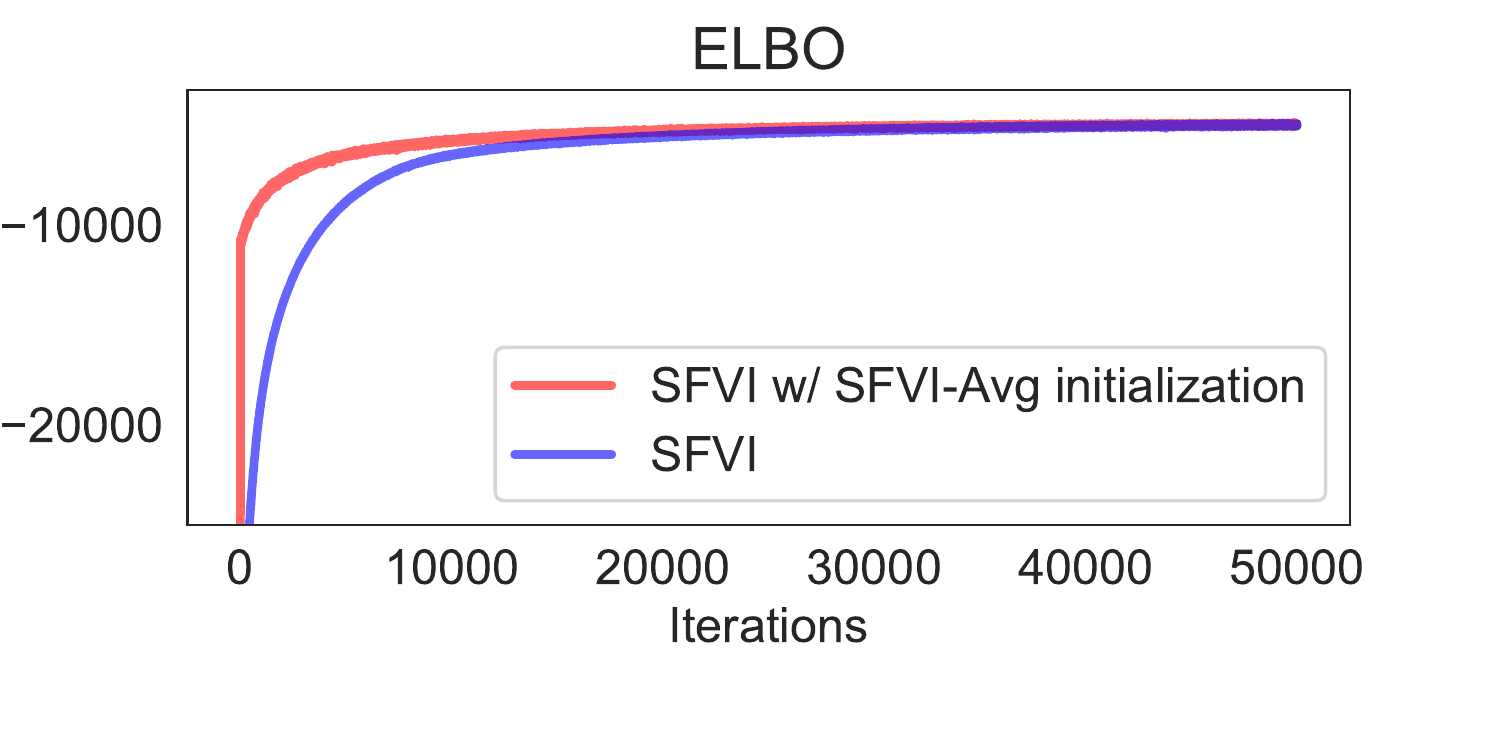}
    \caption{SFVI-Avg initialisation of SFVI rather than from scratch.}
    \label{fig:fast_start}
\end{figure}

The first example considers the MNIST dataset of handwritten digits \citep{lecun1998mnist}, for which each sample is a $28 \times 28$ pixel image (784 pixels total) with an associated label representing the handwritten digit in the image (0--9). The model is
\[\begin{split}
W_{jk} &\sim_{\rm iid} \mathcal{N}(0, \sigma_{\m W}^2), \quad j =1,\ldots, 10, \ k = 1,\ldots,784 \\ 
 b_j &\sim_{\rm iid} \mathcal{N}(\v 0, \sigma^2_{b}), \quad j=1,\ldots,10,  \\  
\v c_k | \m W, \v b &\sim_{\rm iid} {\rm Multinomial}(10, {\rm logit}^{-1}(\m W \v x_k  + \v b) ), 
\end{split} 
\]
for $k=1,\ldots, n$, where $n$ is the total number of data points, and $\sigma_{\m W}^2$ and $\sigma_b^2$ are positive-valued hyperparameters that are learned during training. 
 Thus, for this example $\v Z_G = ({\rm vec}({\m W}), \v b^\top)^\top$, $\v Z_L = \emptyset$, and $\v \theta = (\sigma_{W}^2, \sigma_b^{2})$. This is an {\em empirical Bayes} inference procedure (see \citet[Ch. 3.9]{pml2Book}) as we are optimizing the hyperparameters of the prior distributions of the latent variables. The data are split up evenly across silos. Due to the high-dimensionality of $\v Z_G$ ($n_G = 7850$), an independence approximation is used (i.e., $\m L_G \equiv \m I)$ to enable efficient barycenter computation for {\em SFVI-Avg} via the analytical solution. Table \ref{table:MNIST1} reports the mean posterior-predictive accuracy on the train and test set. Worth noting is that using {\em SFVI-Avg} a number of times during training seems to exhibit a non-trivial effect on generalization performance (when compared to simply averaging once). Additionally, we found in conducting experiments in this paper that using SFVI-Avg for as little as five communication rounds with $10^3$ local steps to initialize default SFVI can reduce the number of steps required to reach convergence by a factor of two, as shown in Figure \ref{fig:fast_start}. 

\begin{table}[ht]
\caption{Results for Multinomial Regression on the MNIST dataset.} \label{table:MNIST1}
\centering
\begin{tabular}{c c c c c}
&        & \multicolumn{2}{c}{\textbf{Accuracy} (\%)}                      \\ 
$J$ \text{ } $N_j$ & \textbf{Method}        & Train         & Test         & \textbf{Rounds }             \\ \hline
25\text{ } 200 & Independent   &      73.0      & \textcolor{red}{\textbf{51.5}}      & 0                   \\
 & SFVI-Avg$(5\cdot 10^4)$ &     \textcolor{red}{\textbf{65.7}}        & 63.9        & 1                  \\
 & SFVI-Avg($10^3$)  &    71.3          & 69.3        &  50                   \\
& SFVI     &      83.8         & \textbf{84.0}        & $5\cdot 10^4$ \\ \hline
5 \text{ } $10^4$ & Independent   &      88.7         & 86.8        & 0                   \\
& SFVI-Avg$(5\cdot 10^4)$ &     90.4         & 90.4         & 1                  \\
& SFVI-Avg($10^3$)  &     90.1          & 90.2        &  50                   \\
& SFVI     &     \textbf{90.7}         &  90.8 & $5\cdot 10^4$ \\ \hline
\end{tabular}
\end{table}
\newpage
% \bibliographystyle{abbrvnat}
% \bibliography{references}

\end{document}